\documentclass{article}

\usepackage{arxiv}
\usepackage{cite}
\usepackage{amsmath,amssymb,amsfonts}
\usepackage{algorithmic}
\usepackage{graphicx}
\usepackage{textcomp}
\usepackage{listings}
\usepackage{booktabs}
\usepackage{arxiv}

\usepackage[utf8]{inputenc} 
\usepackage[T1]{fontenc}    
\usepackage{hyperref}       
\usepackage{url}            
\usepackage{booktabs}       
\usepackage{amsfonts}       
\usepackage{nicefrac}       
\usepackage{float}
\usepackage{microtype}      
\usepackage{lipsum}
\usepackage{graphicx}
\graphicspath{ {./images/} }

\title{Benchmarking Zero-Shot Reasoning Approaches for Error Detection in Solidity Smart Contracts}

\author{
\uppercase{}\authorrefmark{1},
\uppercase{}\authorrefmark{2},
\uppercase{}\authorrefmark{1}, and
\uppercase{}\authorrefmark{1}
}

\author{
 Eduardo Sardenberg \\
  TeleM\'idia Lab - PUC-Rio\\
  Rio de Janeiro, RJ - Brazil \\
  \texttt{etavares@inf.puc-rio.br} \\
   \And
 Antonio Jos\'e G. Busson  \\
  BTG Pactual \\
  São Paulo, SP - Brazil \\
  \texttt{antonio.busson@btgpactual.com} \\
  \And
 Daniel Moraes \\
  TeleM\'idia Lab - PUC-Rio\\
  Rio de Janeiro, RJ - Brazil \\
  \texttt{dmoraes@inf.puc-rio.br} \\
    \And
 Julio Cesar Duarte \\
  Instituto Militar de Engenharia\\
  Rio de Janeiro, RJ - Brazil \\
  \texttt{duarte@ime.eb.br} \\
    \And
 S\'ergio Colcher \\
  TeleM\'idia Lab - PUC-Rio\\
  Rio de Janeiro, RJ - Brazil \\
  \texttt{colcher@inf.puc-rio.br} \\
}

\begin{document}
\maketitle
\begin{abstract}
Smart contracts play a central role in blockchain systems by encoding financial and operational logic. Still, their susceptibility to subtle security flaws poses significant risks of financial loss and erosion of trust. LLMs create new opportunities for automating vulnerability detection, yet the effectiveness of different prompting strategies and model choices in real-world contexts remains uncertain. This paper evaluates state-of-the-art LLMs on Solidity smart contract analysis using a balanced dataset of 400 contracts under two tasks: (i) \emph{Error Detection}, where the model performs binary classification to decide whether a contract is vulnerable, and (ii) \emph{Error Classification}, where the model must assign the predicted issue to a specific vulnerability category. Models are evaluated using zero-shot prompting strategies, including zero-shot, zero-shot Chain-of-Thought (CoT), and zero-shot Tree-of-Thought (ToT). In the Error Detection task, CoT and ToT substantially increase recall (often approaching $\approx 95$--$99\%$), but typically reduce precision, indicating a more sensitive decision regime with more false positives. In the Error Classification task, Claude 3 Opus attains the best Weighted F1-score (90.8) under the ToT prompt, followed closely by its CoT.
\end{abstract}

\maketitle

\section{Introduction}
\label{sec:intro}

Smart contracts are self-executing programs deployed on blockchain networks that automatically enforce the terms of an agreement once predefined conditions are met~\cite{b4}. By eliminating intermediaries, they offer efficiency, transparency, and trust in decentralized applications~\cite{b5}. However, even minor vulnerabilities in smart contract code can lead to irreversible financial losses, security breaches, and systemic failures across blockchain ecosystems. Thus, ensuring the correctness, security, and robustness of these contracts is a critical challenge in secure software engineering.

Traditional approaches to smart contract verification, such as formal methods, static analysis, and manual audits, require specialized expertise and significant human effort. Recently, LLMs (Large Language Models) have emerged as a promising alternative for automating software engineering tasks, including code generation, repair, summarization, and vulnerability detection~\cite{b33, b1, b2}. These models, when paired with well-crafted prompts, can perform complex reasoning over code without explicit task-specific training.

Prior studies have shown that prompt engineering can significantly influence the performance of LLMs in vulnerability detection. For example,   Chen et al.\cite{b18} demonstrated how prompt design improves the detection of common smart contract bugs. Xiao et al.\cite{b28} evaluated LLMs across Solidity versions and found that although prompt tuning can reduce false positives, many models struggle to adapt to updated syntax and vulnerability patterns introduced in recent compiler versions (e.g., Solidity v0.8).

Despite these advances, key questions remain. Most existing work focuses on few-shot prompting or handcrafted templates, often overlooking the potential of reasoning-driven, zero-shot strategies, which require models to detect vulnerabilities without access to labeled examples or manual cues. Furthermore, the comparative effectiveness of different state-of-the-art LLMs (e.g., GPT-4, GPT-4o, Gemini, Claude 3) under these prompting strategies remains underexplored, particularly in terms of both detection accuracy and the quality of vulnerability explanations they can provide.

In this paper, we present an investigation into the use of zero-shot prompting strategies. Specifically, reasoning-based techniques such as CoT (Chain-of-Thought)~\cite{b31} and ToT (Tree-of-Thought)~\cite{b32} are used to guide LLMs in analyzing Solidity smart contracts. Our study evaluates two tasks: (i) \emph{Error Detection}, which consists of detecting whether a contract contains an error (binary vulnerability detection), and (ii) \emph{Error Classification}, which consists of classifying the type of error by assigning it to a specific vulnerability category.

The remainder of this paper is structured as follows. Section~\ref{sec:related_work} discusses related work. Section~\ref{sec:method} describes our methodology, including prompt design, model selection, and evaluation protocol. Section~\ref{sec:evaluation} presents our empirical findings. Finally, Section~\ref{sec:conclusion} outlines our conclusions and directions for future work.
\section{Related Work}
\label{sec:related_work}

The intersection of artificial intelligence and software engineering has driven significant advances in program analysis, automated bug detection, and software repair over the past decade~\cite{b3,b6,b7,b8}. Early work in automated program repair, such as GenProg~\cite{b6} and the learning-based approaches of Kim et al.~\cite{b7}, demonstrated the potential of data-driven techniques to generate human-like patches and localize faults efficiently. Simultaneously, program analysis methods, including static, dynamic, and formal verification techniques, provided complementary guarantees around correctness and reliability, although each came with trade-offs between precision and scalability~\cite{b26}.

With the proliferation of machine learning for code, researchers have begun to explore how models can leverage the statistical properties of source code for tasks such as code completion, summarization, bug detection, and fault localization~\cite{b3,b8}. Tools like DeepBugs~\cite{b8} showed that neural models can learn to detect subtle semantic issues, including naming inconsistencies and common programming mistakes, often outperforming hand-crafted heuristics. More recently, pretrained models for code, such as CodeBERT~\cite{b10}, have expanded the reach of language models to a wide variety of software engineering tasks, providing unified representations for both programming and natural languages.

The arrival of large language models (LLMs), such as GPT, Codex, and their derivatives, has further accelerated progress in code intelligence. Recent systematic evaluations have benchmarked LLMs across code generation, completion, repair, and summarization~\cite{b9,b11,b12}.
Chen et al. ~\cite{b9} and ~\cite{b11} highlight that while LLMs often match or surpass specialized tools on mainstream benchmarks, they may still suffer from brittle performance on out-of-distribution tasks, ambiguous specifications, and security-critical settings. 
 Pearce et al  ~\cite{b12} specifically probes the capability of LLMs to identify code vulnerabilities, finding that performance varies greatly depending on the model and the complexity of the security flaw.

Within the context of blockchain and smart contract security, a growing body of work has examined automated vulnerability detection, often using taxonomies that classify analysis tools into static, dynamic, and formal methods~\cite{b26,b14}. Static analysis tools offer scalability and speed but can miss complex bugs or generate false positives, while formal verification ensures correctness at the cost of practical deployment challenges. Dynamic analysis complements these with runtime guarantees, but coverage remains a challenge~\cite{b26}. Sürücü et al.~\cite{b27} review the emergence of data-driven and machine learning-based approaches for smart contract analysis, noting both the promise of learning from real-world datasets and the difficulties of generalizing across contract versions, domains, and evolving vulnerability patterns.

Recent advances have shifted attention toward leveraging LLMs for smart contract vulnerability detection. Chen et al.\cite{b18} explore how prompt engineering influences the ability of LLMs to detect vulnerabilities in Solidity contracts, demonstrating that the design of prompts and reasoning strategies (e.g., chain-of-thought, tree-of-thought) can substantially affect both precision and recall. Xiao et al.~\cite{b28} extend this work by evaluating a variety of LLMs across multiple Solidity versions, uncovering model-specific strengths and weaknesses and revealing the fragility of current solutions when faced with novel or obfuscated bugs. Yu et al.~\cite{b19} move beyond binary detection, introducing frameworks for automated vulnerability explanation and qualitative assessment, aligning with a broader shift in the community toward explainable and assistive AI for secure development workflows.

Ma et al.\cite{b34} present iAudit, which combines domain-specific fine-tuning with a multi-agent architecture for smart contract auditing. The framework employs a two-stage process: a fine-tuned ``Detector'' model first identifies potential vulnerabilities, and a separate ``Reasoner'' model then generates justifications for these vulnerabilities. To refine the output, LLM-based ``Ranker'' and ``Critic'' agents debate the generated reasons to select the most accurate one. This approach significantly outperformed both prompt-based LLMs and other fine-tuned models, achieving an F1-score of over 91\% and a consistency of 38\% in explanations on a curated dataset, demonstrating the power of dedicated, multi-component AI systems for security tasks.

The emphasis on explainability and human-in-the-loop scenarios has become increasingly important as LLM-based analysis tools are integrated into real-world software engineering pipelines~\cite{b20}. Techniques adapted from spectrum-based fault localization and semantic code analysis~\cite{b8,b6} have laid the groundwork for models that not only detect, but also justify and contextualize their predictions for developers. Tann et al.  ~\cite{b13} demonstrate the importance of semantic-aware analysis for pinpointing and understanding vulnerabilities in smart contracts, while Wang et al .~\cite{b14} provide an up-to-date overview of the progress and challenges in automated detection.

Despite this progress, several open questions remain about the robustness, generalizability, and practical deployment of LLMs for software security tasks. Our work builds on this foundation by systematically evaluating multiple LLMs and prompting strategies—ranging from zero-shot to explicit multi-step reasoning—under a unified experimental setup for smart contract vulnerability detection. In addition to measuring detection accuracy, we assess each model's ability to identify and explain specific vulnerabilities, aiming to inform the design of LLM-assisted tools for secure and interpretable smart contract analysis.

\section{Method}
\label{sec:method}

To evaluate the effectiveness of LLMs in detecting vulnerabilities in smart contracts, we conducted an empirical study involving a diverse set of models and reasoning strategies. Our goal is to assess how architectural differences and reasoning capabilities influence model performance in identifying and characterizing security issues in Solidity code.

We selected multiple LLMs spanning commercial, lightweight, and open-source offerings, including GPT-5.2 and GPT-4.1 (along with their mini and nano variants) from OpenAI, as well as O1, O3-mini, and O4-mini. We also included Google’s Gemini 1.5 and 2.0, as well as the Claude 3 family models Sonnet and Haiku from Anthropic. These models represent a broad spectrum of training scales, deployment configurations, and provider ecosystems, enabling us to explore trade-offs between reasoning quality, inference cost, and practical applicability.

To ensure a rigorous evaluation, we used curated datasets comprising Solidity smart contracts with and without known vulnerabilities. Each contract is annotated with ground-truth labels, vulnerability locations, and vulnerability categories, enabling us to evaluate models in terms of vulnerability detection and, if detected, their ability to classify the type of issue accurately. This setup supports both quantitative and qualitative analyses of model outputs.

In addition to comparing model families, we investigated the effect of various prompting strategies on detection performance. We applied three distinct zero-shot reasoning techniques uniformly across all models and inputs: Zero-Shot, Zero-shot CoT, and Zero-shot ToT. This design enables a controlled comparison of how reasoning style influences detection precision, recall, and the quality of explanations.

All models were evaluated under consistent runtime settings, using standardized prompt templates. The source code, prompts, and datasets used in this study are publicly available in our Git repository\footnote{https://github.com/eduardosardenberg/Zero-Shot-Reasoning-for-Error-Detection-in-Solidity-Smart-Contracts} to support reproducibility and further research.

The remainder of this section is structured as follows. Section~\ref{sec:reasoning} describes the reasoning strategies used for prompting. Section~\ref{sec:dataset} presents the datasets and annotations used in our evaluation.

\subsection{Reasoning}
\label{sec:reasoning}

We compare three zero-shot prompting strategies to isolate how explicit reasoning affects both detection performance and explanation quality.

\begin{itemize}
\item \textbf{Zero-Shot:} the model receives a direct instruction to decide whether a contract is vulnerable. The prompt does not request intermediate reasoning, providing a baseline that reflects common ``single-shot'' use in practice.

\item \textbf{Zero-Shot Chain-of-Thought (CoT):} the prompt explicitly requests step-by-step reasoning before the final decision. This setting tests whether decomposing the analysis into intermediate steps improves recall or vulnerability-type attribution, at the possible cost of longer outputs.

\item \textbf{Zero-Shot Tree-of-Thought (ToT):} the prompt asks the model to explore multiple plausible hypotheses (e.g., different vulnerability candidates) and then select the most likely conclusion. We operationalize ToT via instructions to enumerate alternatives and justify the final choice, approximating structured exploration in a single model response.
\end{itemize}

These strategies are motivated by work showing that structured reasoning and planning prompts can change model reliability in complex tasks~\cite{b29}. Importantly, we keep the task and scoring identical across strategies, so that differences can be attributed to prompting rather than changes in available information.

\subsection{Dataset}
\label{sec:dataset}

A key requirement for evaluating vulnerability detection is to control for dataset imbalance and heterogeneous labeling practices, which are known to affect reported results in smart contract analysis benchmarks~\cite{b16,b26}. We therefore built a dataset of 400 Solidity smart contracts, balanced between vulnerable (200) and non-vulnerable (200) samples. As summarized in Table~\ref{tab:dataset_stats}, contracts were collected from five widely used public sources covering both vulnerability corpora and production-grade code: DAppSCAN~\cite{b22,b15}, JiuZhou~\cite{b21}, Uniswap V2~\cite{b23}, Uniswap V3~\cite{b24}, and OpenZeppelin~\cite{b25}.

\begin{table}[h]
\centering
\caption{Dataset composition and vulnerability distribution}
\label{tab:dataset_stats}
\begin{tabular}{lccc}
\toprule
\textbf{Dataset} & \textbf{Total Files} & \textbf{Error Files} & \textbf{Error Rate (\%)} \\
\midrule
DAppSCAN & 200 & 200 & 100.0 \\
JiuZhou & 102 & 0 & 0.0 \\
UniswapV2 & 12 & 0 & 0.0 \\
UniswapV3 & 62 & 0 & 0.0 \\
openZeppelin & 24 & 0 & 0.0 \\
\bottomrule
\end{tabular}
\end{table}

\paragraph{Vulnerable contracts.}
We selected 200 vulnerable contracts from DAppSCAN~\cite{b22,b15}, which aggregates real-world weaknesses observed in dApp projects. Each vulnerable contract is associated with a weakness identifier following the Smart Contract Weakness Classification (SWC) scheme, enabling both binary detection evaluation and vulnerability-type attribution. The distribution of SWC categories in our vulnerable subset is shown in Table~\ref{tab:swc_distribution}, spanning common classes such as reentrancy, unchecked return values, integer bugs, transaction-order dependence, and insecure compiler/version patterns.

\paragraph{Non-vulnerable contracts.}
For the non-vulnerable set, we sampled contracts from widely used, open-source deployments and reference libraries, which are commonly used as ``benign'' baselines in empirical smart contract studies~\cite{b16}. Specifically, we included 102 contracts from JiuZhou~\cite{b21}, 62 from Uniswap V3~\cite{b24}, 12 from Uniswap V2~\cite{b23}, and 24 from OpenZeppelin~\cite{b25}. These sources are representative of the type of Solidity code found in production and widely reused across the ecosystem.

\paragraph{Data format and annotations.}
To support scalable evaluation and reproducibility, we standardized all samples into a single CSV file. Each entry contains (i) the dataset origin (\texttt{dataset\_name}), (ii) source location (\texttt{file\_path} and \texttt{file\_name}), (iii) semantic grouping (\texttt{category}), and (iv) the ground-truth label (\texttt{has\_error}). For vulnerable contracts, we additionally keep the vulnerability category used to compute per-class performance and to evaluate whether model explanations match the intended weakness type. This structure enables automated prompt generation and consistent scoring across models and prompting strategies.

\begin{table}[H]
\centering
\caption{Distribution of SWC vulnerability categories in the DAppSCAN dataset.}
\label{tab:swc_distribution}
\begin{tabular}{p{5cm}rr}
\toprule
\textbf{SWC Category} & \textbf{Count} & \textbf{Percent (\%)} \\
\midrule
SWC-101-Integer Overflow and Underflow & 70 & 18.23 \\
SWC-135-Code With No Effects & 64 & 16.67 \\
SWC-104-Unchecked Call Return Value & 39 & 10.16 \\
SWC-114-Transaction Order Dependence & 25 & 6.51 \\
SWC-102-Outdated Compiler Version & 25 & 6.51 \\
SWC-128-DoS With Block Gas Limit & 23 & 5.99 \\
SWC-107-Reentrancy & 22 & 5.73 \\
SWC-103-Floating Pragma & 21 & 5.47 \\
SWC-116-Block values as a proxy for time & 15 & 3.91 \\
SWC-100-Function Default Visibility & 15 & 3.91 \\
SWC-105-Unprotected Ether Withdrawal & 11 & 2.86 \\
SWC-108-State Variable Default Visibility & 8 & 2.08 \\
SWC-112-Delegatecall to Untrusted Callee & 6 & 1.56 \\
SWC-115-Authorization through tx.origin & 5 & 1.30 \\
SWC-119-Shadowing State Variables & 5 & 1.30 \\
SWC-113-DoS with Failed Call & 5 & 1.30 \\
SWC-129-Typographical Error & 4 & 1.04 \\
SWC-131-Presence of unused variables & 3 & 0.78 \\
SWC-124-Write to Arbitrary Storage Location & 3 & 0.78 \\
SWC-110-Assert Violation & 2 & 0.52 \\
SWC-122-Lack of Proper Signature Verification & 2 & 0.52 \\
SWC-111-Use of Deprecated Solidity Functions & 2 & 0.52 \\
SWC-126-Insufficient Gas Griefing & 2 & 0.52 \\
SWC-117-Signature Malleability & 1 & 0.26 \\
SWC-125-Incorrect Inheritance Order & 1 & 0.26 \\
SWC-123-Requirement Violation & 1 & 0.26 \\
SWC-106-Unprotected SELFDESTRUCT Instruction & 1 & 0.26 \\
SWC-121-Missing Protection against Signature Replay Attacks & 1 & 0.26 \\
SWC-133-Hash Collisions With Multiple Variable Length Arguments & 1 & 0.26 \\
SWC-120-Weak Sources of Randomness from Chain Attributes & 1 & 0.26 \\
\bottomrule
\end{tabular}
\end{table}

\section{Experiments}
\label{sec:evaluation}

We report results for two complementary tasks. First, in the \emph{Error Detection} task, the model performs binary classification by deciding whether a contract is vulnerable, which captures the ability to flag risky code while balancing false alarms. Second, in the \emph{Error Classification} task, we assess whether the model can correctly attribute the predicted issue to a specific vulnerability category, which better reflects practical auditing scenarios where developers need both a warning and an actionable label. Throughout, we summarize performance using precision, recall, and F1-score, and we discuss the trade-offs introduced by different prompting strategies in Section~\ref{sec:discussion}.

\subsection{Error Detection Task}
\label{sec:error_detection_task}

\begin{table}[H]
\centering
\scriptsize
\caption{Error detection (binary vulnerability classification) results across LLMs and zero-shot prompting strategies sorted by F1-score}
\label{tab:experiment1}
\begin{tabular}{lcccc}
\toprule
\textbf{LLM} & \textbf{Prompt} & \textbf{Precision} & \textbf{Recall} & \textbf{F1-Score} \\
\midrule

gpt-4.1              & zero-shot         & 70.12 & 89.73 & 78.83 \\
gpt-4.1              & zero-shot-cot     & 65.32 & 98.07 & 78.79 \\
gpt-4.1              & zero-shot-tot     & 64.10 & 99.08 & 78.06 \\
Claude 3 Sonnet      & zero-shot-cot     & 69.90 & 87.13 & 77.47 \\
gpt-5                & zero-shot-tot     & 62.56 & 98.57 & 76.61 \\ 
Claude 3 Sonnet      & zero-shot-tot     & 69.74 & 83.62 & 76.10 \\
Claude 4.2 Opus      & zero-shot-tot     & 63.21 & 92.14 & 75.22 \\
Claude 4.2 Opus      & zero-shot-cot     & 62.88 & 92.37 & 75.08 \\
gpt-4.1-mini         & zero-shot         & 65.78 & 86.47 & 74.71 \\
Claude 4.2 Opus      & zero-shot         & 62.13 & 93.42 & 74.53 \\ 
gpt-5.2              & zero-shot-tot     & 61.97 & 93.12 & 74.50 \\ 
gpt-4o               & zero-shot         & 64.81 & 87.43 & 74.44 \\
gpt-4.1-mini         & zero-shot-cot     & 60.06 & 97.00 & 74.19 \\
gpt-4.1-mini         & zero-shot-tot     & 58.92 & 98.43 & 73.46 \\
Claude 3 Opus        & zero-shot         & 64.86 & 84.17 & 73.32 \\
gemini-1.5-flash     & zero-shot-cot     & 58.28 & 98.47 & 73.28 \\
gemini-2.0-flash     & zero-shot         & 59.94 & 93.44 & 73.01 \\
gemini-2.0-flash     & zero-shot-tot     & 57.94 & 98.44 & 73.01 \\
gpt-5.2              & zero-shot-cot     & 59.43 & 94.47 & 72.98 \\
gpt-5                & zero-shot         & 56.82 & 99.87 & 72.50 \\ 
gemini-2.0-flash     & zero-shot-cot     & 58.51 & 94.42 & 72.26 \\
gemini-1.5-flash     & zero-shot         & 57.18 & 97.43 & 72.06 \\
gemini-1.5-flash     & zero-shot-tot     & 57.40 & 97.12 & 72.18 \\
gpt-4o               & zero-shot-cot     & 54.87 & 98.44 & 70.53 \\
gpt-5.2              & zero-shot         & 54.35 & 99.06 & 70.25 \\ 
gpt-4o-mini          & zero-shot-tot     & 53.95 & 99.04 & 70.20 \\
gpt-5                & zero-shot-cot     & 54.76 & 97.43 & 70.18 \\ 
gpt-4o-mini          & zero-shot-cot     & 54.42 & 98.47 & 70.17 \\
gpt-4o               & zero-shot-tot     & 54.27 & 98.43 & 70.05 \\
gpt-4o-mini          & zero-shot         & 56.56 & 90.43 & 69.59 \\
Claude 3 Opus        & zero-shot-cot     & 65.40 & 69.83 & 67.54 \\
gpt-4.1-nano         & zero-shot-cot     & 51.80 & 95.73 & 67.31 \\
gpt-4.1-nano         & zero-shot-tot     & 51.02 & 98.17 & 67.25 \\
Claude 3 Sonnet      & zero-shot         & 67.51 & 66.43 & 66.96 \\
o1                   & zero-shot-cot     & 50.81 & 95.73 & 66.46 \\
o1                   & zero-shot-tot     & 49.32 & 98.10 & 65.61 \\
gpt-4.1-nano         & zero-shot         & 55.12 & 81.25 & 65.53 \\
o1                   & zero-shot         & 52.03 & 81.25 & 63.46 \\
o3-mini              & zero-shot-tot     & 46.31 & 97.15 & 62.52 \\
Claude 3 Opus        & zero-shot-tot     & 64.82 & 59.41 & 62.02 \\
o3-mini              & zero-shot-cot     & 46.82 & 91.23 & 61.74 \\
o4-mini              & zero-shot-cot    & 44.39 & 89.15 & 59.01 \\
o3-mini              & zero-shot        & 48.10 & 76.92 & 58.97 \\
o4-mini              & zero-shot-tot    & 43.90 & 95.28 & 58.97 \\
o4-mini              & zero-shot        & 45.85 & 75.47 & 56.87 \\
Claude 3 Haiku       & zero-shot        & 68.42 & 45.73 & 54.82 \\
Claude 3 Haiku       & zero-shot-tot    & 44.27 & 15.91 & 23.18 \\
Claude 3 Haiku       & zero-shot-cot    & 46.51 & 10.26 & 16.81 \\
\bottomrule
\end{tabular}
\end{table}

Table~\ref{tab:experiment1} reports the binary vulnerability detection results for each model under the three prompting strategies. Overall, the best F1-scores are obtained by large, general-purpose models (e.g., GPT-4.1 and Claude Sonnet/Opus variants), while smaller models (mini/nano tiers) exhibit noticeably lower precision and/or recall. This indicates that model capacity remains a key factor for detecting subtle security issues that require reasoning over control flow and external-call semantics.

Across most model families, CoT and ToT prompts substantially increase recall (often approaching $\approx 95$--$99\%$), but typically reduce precision. In practice, explicit reasoning makes models more ``sensitive'': they flag a larger fraction of truly vulnerable contracts, but also produce more false positives. In contrast, the plain zero-shot prompt is usually more conservative, frequently yielding higher precision but lower recall, which can be preferable in settings where false alarms are costly.

Prompt sensitivity is not uniform across model versions: some strong models benefit primarily from the zero-shot baseline (e.g., GPT-4.1 achieves its highest F1 in zero-shot), while others peak under explicit reasoning (e.g., Claude 3 Sonnet under CoT and GPT-5 under ToT). Notably, GPT-5 benefits most from ToT in our setting, which shifts it toward a recall-oriented operating point, consistent with the precision--recall trade-off induced by reasoning-based prompts. This reinforces that prompt selection should be treated as a model-specific hyperparameter rather than a universally beneficial recipe.

Smaller models are more fragile under multi-step prompting. Nano/mini configurations and compact reasoning-oriented models (e.g., o3-mini and o4-mini) often show large recall gains under CoT/ToT, but with enough precision degradation to offset those gains, resulting in lower F1 than larger counterparts. An extreme case is Claude 3 Haiku, where CoT sharply reduces recall, suggesting that additional reasoning instructions can destabilize behavior in limited-capacity models.

In summary, binary detection highlights a consistent precision--recall trade-off: structured reasoning prompts increase coverage (recall) but may introduce additional false positives (precision loss). The net effect depends strongly on the model family/version and model size, so practitioners should select prompts to match the desired operating point (high-recall screening vs. high-precision triage).

\subsection{Error Classification Task}
\label{sec:eval_results}

\begin{table}H]
\centering
\caption{Multi-class error classification performance across LLMs and zero-shot prompting strategies, ordered by Weighted F1-score}
\label{tab:weighted_f1_sorted}
\scriptsize
\begin{tabular}{l l c c c c c c c c c}
\hline
\textbf{LLM} & \textbf{Prompt} &
\multicolumn{3}{c}{\textbf{Micro}} &
\multicolumn{3}{c}{\textbf{Macro}} &
\multicolumn{3}{c}{\textbf{Weighted}} \\
 &  & P & R & F1 & P & R & F1 & P & R & F1 \\
\hline

Claude 3 Opus & zero-shot-tot & 89.7 & 89.7 & 89.7 & 78.1 & 78.1 & 77.1 & 94.8 & 89.7 & \textbf{90.8} \\
Claude 3 Opus & zero-shot-cot & 90.0 & 90.0 & 90.0 & 84.4 & 87.4 & 84.6 & 91.9 & 90.0 & \textbf{89.9} \\
Claude 3 Opus & zero-shot & 85.5 & 85.5 & 85.5 & 81.3 & 84.2 & 81.7 & 91.4 & 85.5 & \textbf{87.1} \\
Claude 4.6 Opus & zero-shot-tot & 82.4 & 82.4 & 82.4 & 70.3 & 66.8 & 67.1 & 93.1 & 82.4 & \textbf{85.4} \\
gpt-4o & zero-shot-tot & 82.8 & 82.8 & 82.8 & 83.9 & 85.0 & 81.2 & 92.5 & 82.8 & \textbf{83.4} \\
gpt-4o & zero-shot & 82.0 & 82.0 & 82.0 & 68.1 & 75.8 & 69.6 & 82.7 & 82.0 & \textbf{81.3} \\
Claude 4.6 Opus & zero-shot & 80.0 & 80.0 & 80.0 & 67.5 & 73.2 & 68.3 & 83.8 & 80.0 & \textbf{80.5} \\
gpt-4.1-mini & zero-shot-tot & 79.3 & 79.3 & 79.3 & 73.9 & 77.8 & 72.7 & 87.4 & 79.3 & \textbf{80.4} \\
gpt-4o-mini & zero-shot-tot & 79.3 & 79.3 & 79.3 & 73.9 & 77.8 & 72.7 & 87.4 & 79.3 & \textbf{80.4} \\
o1 & zero-shot-cot & 72.6 & 72.6 & 72.6 & 70.4 & 71.8 & 71.1 & 83.5 & 72.6 & \textbf{77.7} \\
gpt-4.1-mini & zero-shot & 72.4 & 72.4 & 72.4 & 72.1 & 66.1 & 63.7 & 91.5 & 72.4 & \textbf{76.2} \\
gpt-4o-mini & zero-shot & 72.4 & 72.4 & 72.4 & 72.1 & 66.1 & 63.7 & 91.5 & 72.4 & \textbf{76.2} \\
o1 & zero-shot-tot & 71.3 & 71.3 & 71.3 & 66.2 & 68.1 & 67.1 & 81.4 & 71.3 & \textbf{76.0} \\
Claude 3 Haiku & zero-shot & 71.5 & 71.5 & 71.5 & 71.8 & 71.5 & 70.4 & 84.5 & 71.5 & \textbf{75.7} \\
gpt-4.1 & zero-shot & 73.0 & 73.0 & 73.0 & 62.5 & 65.4 & 61.0 & 79.5 & 73.0 & \textbf{73.4} \\
Claude 4.6 Opus & zero-shot-cot & 71.5 & 71.5 & 71.5 & 59.3 & 63.0 & 57.6 & 77.5 & 71.5 & \textbf{72.0} \\
o4-mini & zero-shot-cot & 64.2 & 64.2 & 64.2 & 60.5 & 62.3 & 61.4 & 81.7 & 64.2 & \textbf{71.9} \\
gpt-4.1-nano & zero-shot & 69.0 & 69.0 & 69.0 & 62.5 & 57.4 & 57.3 & 84.5 & 69.0 & \textbf{71.6} \\
gemini-2.0-flash & zero-shot & 71.0 & 71.1 & 71.2 & 62.6 & 60.7 & 58.2 & 78.2 & 71.0 & \textbf{71.2} \\
o1 & zero-shot & 69.0 & 69.0 & 69.0 & 67.7 & 68.2 & 65.6 & 78.7 & 69.0 & \textbf{71.2} \\
gemini-2.0-flash & zero-shot-cot & 69.0 & 69.0 & 69.1 & 58.9 & 57.7 & 55.1 & 79.4 & 69.0 & \textbf{71.1} \\
gpt-5 & zero-shot-tot & 67.6 & 67.6 & 67.6 & 61.3 & 60.6 & 56.3 & 86.9 & 67.6 & \textbf{70.0} \\
o4-mini & zero-shot-tot & 62.6 & 62.6 & 62.6 & 55.4 & 57.1 & 56.2 & 79.2 & 62.6 & \textbf{69.9} \\
gemini-2.5-pro & zero-shot & 69.5 & 69.5 & 69.5 & 62.5 & 62.4 & 57.7 & 76.9 & 69.5 & \textbf{69.8} \\
gpt-4.1 & zero-shot-cot & 65.5 & 65.5 & 65.5 & 69.8 & 66.1 & 62.6 & 85.1 & 65.5 & \textbf{68.2} \\
o3-mini & zero-shot-cot & 61.2 & 61.2 & 61.2 & 54.6 & 56.1 & 55.3 & 75.4 & 61.2 & \textbf{67.6} \\
gpt-4.1 & zero-shot-tot & 69.0 & 69.0 & 69.0 & 53.4 & 60.3 & 54.4 & 70.1 & 69.0 & \textbf{67.2} \\
gpt-4.1-mini & zero-shot-cot & 62.1 & 62.1 & 62.1 & 65.5 & 56.2 & 54.0 & 87.8 & 62.1 & \textbf{66.7} \\
gpt-4o-mini & zero-shot-cot & 62.1 & 62.1 & 62.1 & 65.5 & 56.2 & 54.0 & 87.8 & 62.1 & \textbf{66.7} \\
o4-mini & zero-shot & 61.5 & 61.5 & 61.5 & 57.4 & 59.3 & 53.2 & 78.4 & 61.5 & \textbf{65.6} \\
o3-mini & zero-shot & 58.7 & 58.7 & 58.7 & 50.3 & 52.4 & 51.3 & 73.6 & 58.7 & \textbf{65.3} \\
gpt-4o & zero-shot-cot & 65.5 & 65.5 & 65.5 & 53.0 & 55.1 & 50.2 & 74.3 & 65.5 & \textbf{64.9} \\
gemini-2.0-flash & zero-shot-tot & 59.0 & 59.0 & 59.0 & 53.3 & 51.8 & 50.5 & 74.7 & 59.0 & \textbf{64.1} \\
o3-mini & zero-shot-tot & 57.4 & 57.4 & 57.4 & 48.5 & 50.2 & 49.3 & 71.3 & 57.4 & \textbf{63.6} \\
gpt-5 & zero-shot & 59.0 & 59.0 & 59.0 & 51.4 & 60.0 & 50.3 & 69.9 & 59.0 & \textbf{60.5} \\
Claude 3 Haiku & zero-shot-cot & 51.5 & 51.5 & 51.5 & 65.7 & 52.6 & 55.8 & 82.0 & 51.5 & \textbf{58.5} \\
gemini-2.5-pro & zero-shot-cot & 59.0 & 59.0 & 59.0 & 51.1 & 53.5 & 47.3 & 70.3 & 59.0 & \textbf{58.2} \\
gpt-5.2 & zero-shot & 56.5 & 56.5 & 56.5 & 54.3 & 53.2 & 47.4 & 72.0 & 56.5 & \textbf{57.9} \\
gpt-5.2 & zero-shot-tot & 55.9 & 55.9 & 55.9 & 51.1 & 47.2 & 43.2 & 78.0 & 55.9 & \textbf{56.9} \\
gemini-2.5-pro & zero-shot-tot & 56.0 & 56.0 & 56.0 & 42.6 & 46.0 & 39.8 & 62.9 & 56.0 & \textbf{54.2} \\
Claude 3 Haiku & zero-shot-tot & 51.7 & 51.7 & 51.7 & 39.5 & 38.6 & 36.8 & 63.8 & 51.7 & \textbf{53.7} \\
gpt-5 & zero-shot-cot & 50.0 & 50.0 & 50.0 & 43.5 & 41.7 & 36.9 & 72.4 & 50.0 & \textbf{52.0} \\
gpt-5.2 & zero-shot-cot & 52.9 & 52.9 & 52.9 & 35.3 & 39.5 & 35.1 & 54.9 & 52.9 & \textbf{50.7} \\
gpt-4.1-nano & zero-shot-tot & 48.3 & 48.3 & 48.3 & 41.9 & 42.6 & 36.7 & 61.3 & 48.3 & \textbf{47.7} \\
gpt-4.1-nano & zero-shot-cot & 44.8 & 44.8 & 44.8 & 50.5 & 45.1 & 43.0 & 64.9 & 44.8 & \textbf{47.5} \\
gemini-2.5-flash & zero-shot-tot & 44.5 & 44.5 & 44.5 & 43.1 & 43.3 & 37.9 & 58.8 & 44.5 & \textbf{45.8} \\
gemini-2.5-flash & zero-shot & 44.0 & 44.0 & 44.0 & 31.8 & 30.9 & 27.3 & 58.1 & 44.0 & \textbf{44.6} \\
gemini-2.5-flash & zero-shot-cot & 41.5 & 41.5 & 41.5 & 36.1 & 34.5 & 30.7 & 55.5 & 41.5 & \textbf{42.1} \\
\hline
\end{tabular}
\end{table}

\begin{table}[H]
\centering
\caption{Distribution of SWC vulnerability categories \emph{predicted by our best model} (Claude 3 Opus with Tree-of-Thought) for the evaluated subset of 200 smart contracts}
\label{tab:swc_distribution_sorted}
\scriptsize
\begin{tabular}{l r l r}
\hline
\textbf{Class} & \textbf{Count} & \textbf{Class} & \textbf{Count} \\
\hline
SWC-101 & 22 & SWC-105 & 5  \\
SWC-114 & 21 & SWC-108 & 5  \\
SWC-135 & 21 & SWC-113 & 5  \\
SWC-103 & 19 & SWC-115 & 5  \\
SWC-104 & 18 & SWC-112 & 3  \\
SWC-107 & 18 & SWC-119 & 3  \\
SWC-102 & 16 & SWC-120 & 2  \\
SWC-116 & 10 & SWC-131 & 2  \\
SWC-128 & 9  & NONE    & 1  \\
SWC-100 & 7  & SWC-110 & 1  \\
\hline
\end{tabular}
\end{table}

Table~\ref{tab:weighted_f1_sorted} reports multi-class vulnerability classification results, in which the model must both identify whether a contract is vulnerable and assign the correct SWC category (or \texttt{NONE}). We report Micro, Macro, and Weighted precision/recall/F1 to provide complementary views of performance: Micro emphasizes overall accuracy dominated by frequent classes, Macro highlights robustness across all categories (including rare ones), and Weighted balances these perspectives by accounting for class frequency while still penalizing systematic confusion.

The strongest results are achieved by larger, frontier models, confirming that accurate vulnerability \emph{classification} is more demanding than binary detection. In particular, \textbf{Claude 3 Opus} attains the best Weighted F1-score (\textbf{90.8}) under the ToT prompt, with its CoT and zero-shot variants performing closely behind. Other high-capacity models (e.g., GPT-4o and newer Opus variants) remain competitive, whereas compact models (mini/nano tiers and small ``reasoning'' models) generally lag, reflecting limited ability to disambiguate between closely related SWC categories.

Structured reasoning prompts do not uniformly improve multi-class classification. For some models, ToT yields clear gains (e.g., Claude 3 Opus and GPT-5), consistent with the hypothesis that explicitly exploring alternatives helps disambiguate between similar vulnerability hypotheses. For other models, however, CoT/ToT can degrade Macro and Weighted scores (e.g., GPT-4.1), indicating that longer reasoning traces may introduce label drift, over-commitment to plausible-but-wrong categories, or inconsistent use of the SWC taxonomy.

Micro scores primarily reflect performance on frequent classes and on the dominant \texttt{NONE}/common-SWC decisions, whereas Macro scores weight all classes equally and therefore expose difficulties on rare or fine-grained vulnerability types. The persistent gap between Micro and Macro metrics across configurations indicates that high aggregate accuracy can coexist with poor robustness on minority SWC categories. Weighted scores sit between these extremes, but can still partially mask systematic confusion on long-tail classes, especially when errors concentrate in a small set of rare categories.

Smaller models are particularly sensitive in the multi-class setting: even when detection recall is acceptable, they tend to confuse related vulnerability categories and exhibit larger drops in Macro F1. This is consistent with limited ability to maintain a stable mapping from natural-language rationales to the correct SWC label, especially under multi-step prompts.

Overall, the Error Classification task is more demanding than binary detection: the best configurations combine high-capacity models with prompting strategies that help separate similar vulnerability hypotheses. However, gains are model-dependent and should be interpreted through class-sensitive metrics (Macro/Weighted), not only aggregate (Micro) performance.

\begin{figure*}[t]
    \centering
    \includegraphics[width=0.8\textwidth]{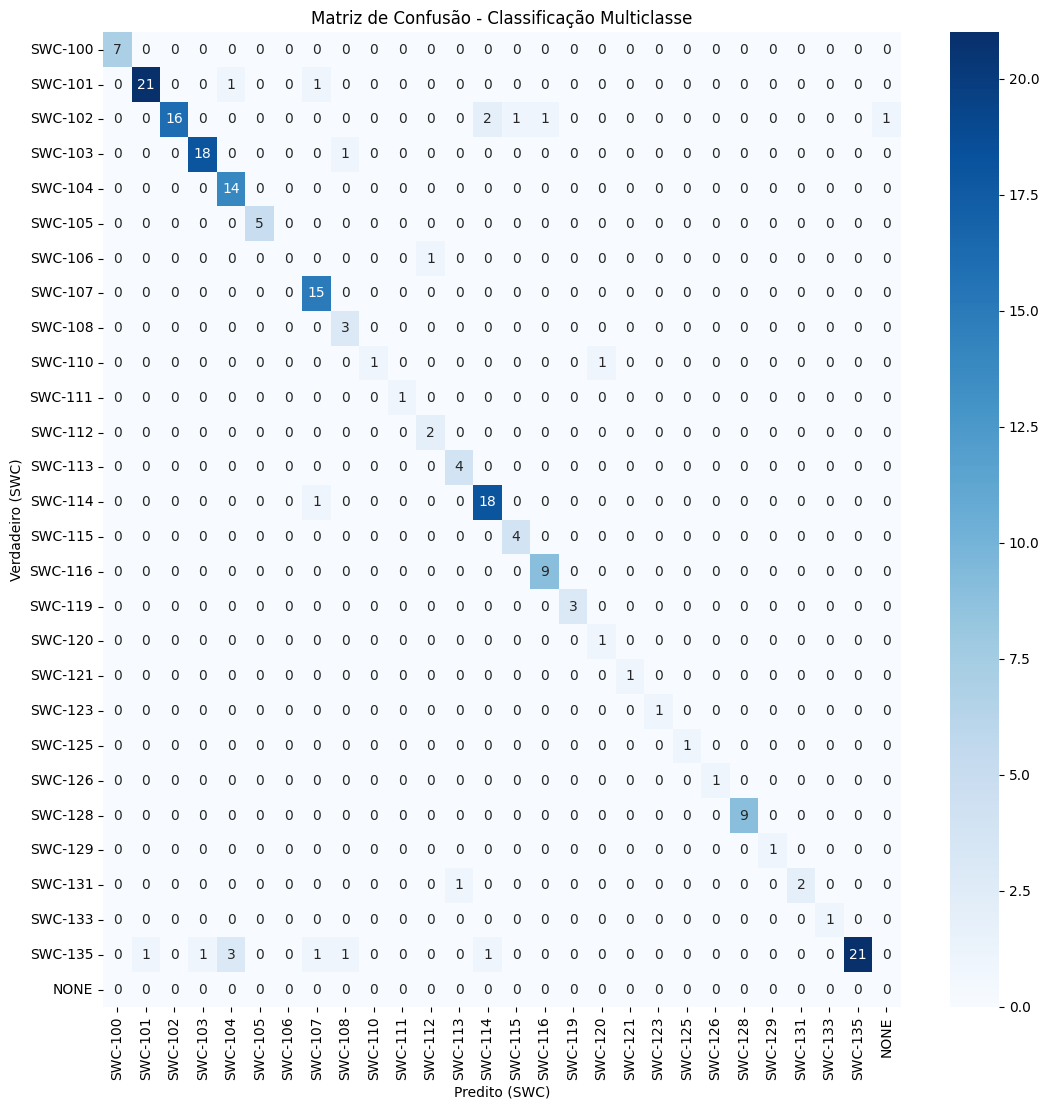}
    \caption{Confusion matrix for the best-performing configuration (Claude 3 Opus with Tree-of-Thought), showing per-class confusions across SWC categories.}
    \label{fig:conf_matrix}
\end{figure*}

Table~\ref{tab:swc_distribution_sorted} presents the distribution of vulnerability classes \emph{predicted by our best model (Claude 3 Opus--ToT)} in the multi-class setting, considering only the subset of 200 smart contracts analyzed in our evaluation, while Figure~\ref{fig:conf_matrix} presents the confusion matrix for the same configuration. Together, they summarize both \emph{what} the model tends to predict (label frequency) and \emph{how} those predictions align with the ground truth (confusions between SWC categories).

\subsection{Discussion}
\label{sec:discussion}

\begin{figure*}[h]
    \centering
    \includegraphics[width=\textwidth]{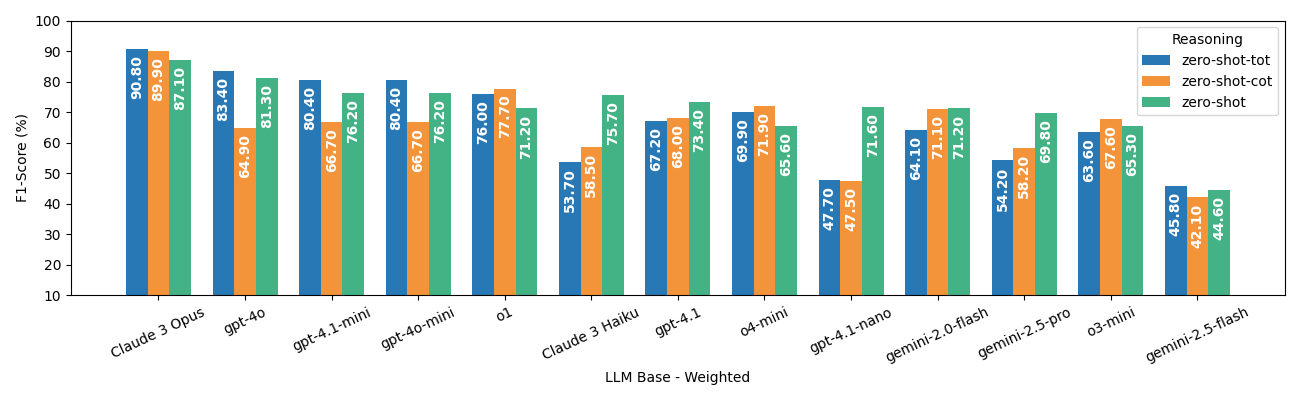}
\caption{Weighted F1-score for the Error Classification task across evaluated LLMs under zero-shot, zero-shot-CoT, and zero-shot-ToT prompting strategies.}
    \label{fig:chart_comp}
\end{figure*}

\begin{table}[H]
\scriptsize
\centering
\caption{F1-score gain of zero-shot-CoT and zero-shot-ToT over the zero-shot baseline in the Error Detection task. Gains are computed as $\Delta\mathrm{F1}=\mathrm{F1}_{\text{prompt}}-\mathrm{F1}_{\text{zero-shot}}$.}
\label{tab:experiment1_gain}
\begin{tabular}{lcc}
\toprule
\textbf{LLM} & \textbf{CoT F1-Score Gain} & \textbf{ToT F1-Score Gain} \\
\midrule
gpt-5.2              & 2.73   & 4.25 \\
gpt-5                & -2.32  & 4.11 \\
gpt-4.1              & -0.04  & -0.77 \\
gpt-4.1-mini         & -0.52  & -1.25 \\
gpt-4.1-nano         & 1.78   & 1.72 \\
gpt-4o               & -3.91  & -4.39 \\
gpt-4o-mini          & 0.58   & 0.61 \\
o4-mini              & 2.14   & 2.10 \\
o3-mini              & 2.77   & 3.55 \\
o1                   & 2.99   & 2.15 \\
gemini-1.5-flash     & 1.22   & 0.12 \\
gemini-2.0-flash     & -0.75  & 0.00 \\
Claude 4.2 Opus      & 0.55   & 0.69 \\
Claude 3 Opus        & -5.78  & -11.30 \\
Claude 3 Haiku       & -38.01 & -31.64 \\
Claude 3 Sonnet      & 10.51  & 9.14 \\
\bottomrule
\end{tabular}
\end{table}
\begin{table}[H]
\centering
\scriptsize
\caption{Weighted F1-score gain of zero-shot-CoT and zero-shot-ToT over the zero-shot baseline in the Error Classification task. Gains are computed as $\Delta\mathrm{F1}_w=\mathrm{F1}_{w,\text{prompt}}-\mathrm{F1}_{w,\text{zero-shot}}$, where $\mathrm{F1}_w$ denotes the Weighted F1-score.}
\label{tab:experiment2_gain}
\begin{tabular}{lcc}
\toprule
\textbf{LLM} & \textbf{CoT F1-Score Gain} & \textbf{ToT F1-Score Gain} \\
\midrule

gpt-5                & -8.5  & 9.5 \\
o1                   & 6.5   & 4.8 \\
Claude 4.6 Opus      & -8.5  & 4.9 \\
o4-mini              & 6.3   & 4.3 \\
gpt-4.1-mini         & -9.5  & 4.2 \\
gpt-4o-mini          & -9.5  & 4.2 \\
Claude 3 Opus        & 2.8   & 3.7 \\
gpt-4o               & -16.4 & 2.1 \\
gemini-2.5-flash     & -2.5  & 1.2 \\
gpt-5.2              & -7.2  & -1.0 \\
o3-mini              & 2.3   & -1.7 \\
gpt-4.1              & -9.5  & -6.2 \\
gemini-2.0-flash     & -0.1  & -7.1 \\
gemini-2.5-pro       & -11.6 & -15.6 \\
Claude 3 Haiku       & -17.2 & -22.0 \\
gpt-4.1-nano         & -24.1 & -23.9 \\
\bottomrule
\end{tabular}
\end{table}

Our results indicate that the impact of reasoning-oriented prompting is highly \emph{model-dependent} and differs across tasks. In the \textbf{Error Detection} task, CoT/ToT prompts typically shift models toward a recall-oriented operating point---often increasing recall at the cost of precision (i.e., more false positives). This pattern is reflected in Table~\ref{tab:experiment1_gain}, where some models benefit from CoT/ToT (positive $\Delta$F1) while others show small or even negative changes.

In the \textbf{Error Classification} task, prompt sensitivity becomes even more pronounced. As shown in Table~\ref{tab:experiment2_gain} and Figure~\ref{fig:chart_comp}, larger, frontier models generally achieve the strongest Weighted F1-scores, but additional reasoning does not uniformly help. While ToT can yield substantial improvements for some models (e.g., GPT-5), CoT/ToT may degrade performance for others, suggesting that longer reasoning traces can introduce label drift, over-commitment to plausible-but-incorrect SWC categories, or inconsistent use of the SWC taxonomy.

Overall, these findings reinforce two practical implications. First, prompt choice (zero-shot vs.\ CoT vs.\ ToT) should be treated as a \emph{model-specific hyperparameter} rather than a universally beneficial recipe. Second, in security-oriented workflows, practitioners must explicitly choose the desired operating point and failure mode: higher recall (broader coverage) versus higher precision (fewer false alarms), and coarse-grained detection versus fine-grained vulnerability attribution.

An illustrative example of the limitations of reasoning-based prompting is presented in the listing below, which shows a Tree-of-Thought (ToT) explanation generated by the \textbf{Claude 3 Opus} model. Here, the model correctly predicts that the contract contains an error; however, the initial reasoning path leads to an incorrect vulnerability label. This demonstrates that a correct high-level decision (``bug'' vs.\ ``no bug'') does not necessarily imply correct vulnerability attribution.

Notably, in a subsequent reasoning attempt, the same model revises its analysis and identifies the underlying issue. This behavior highlights both a strength and a limitation of advanced reasoning strategies: multi-step reasoning can enable self-correction, but intermediate paths may still contain misleading or partially incorrect explanations. Such instability emphasizes the importance of evaluating not only the final prediction, but also the reliability of the reasoning trajectories produced by LLMs, especially in security-critical domains such as smart contract analysis.

\lstset{
  basicstyle=\ttfamily\scriptsize,
  keywordstyle=\color{blue},
  commentstyle=\color{gray},
  stringstyle=\color{teal},
  frame=single,
  breaklines=true,          
  breakatwhitespace=true,   
}

\begin{lstlisting}[]
{
  "file_name": "BaseParaSwapAdapter.sol",
  "has_error_true_label": true,
  "has_error_prediction": true,
  "error_type": "['SWC-107-Reentrancy']",
  "location": 104,
  "reasoning_paths": [
        {
            "path": "Unchecked return value from LENDING_POOL.withdraw",
            "analysis": "The return value of LENDING_POOL.withdraw is not checked, which could lead to unexpected behavior."
        },
        {
            "path": "Potential reentrancy vulnerability in _pullAToken",
            "analysis": "The external call to LENDING_POOL.withdraw is made after the token transfer, which could enable reentrancy attacks."
        },
        {
            "path": "No checks on permit signature validity",
            "analysis": "The _usePermit function does not thoroughly validate the permit signature, which could allow invalid signatures."
        },
        {
            "path": "rescueTokens can be used to drain funds",
            "analysis": "The rescueTokens function allows the contract owner to transfer out any tokens, which could be abused."
        },
        {
            "path": "No checks on pulled aToken amount",
            "analysis": "The _pullAToken function does not validate that the amount of aTokens pulled matches the expected amount."
        }
    ],
    "final_reasoning": "The contract has a potential reentrancy vulnerability and does not properly check the return value of an external call. The unchecked call return value is the most direct vulnerability.",
    "bug_type": "SWC-104-Unchecked Call Return Value"
}
\end{lstlisting}

A contrasting case is illustrated in the next listing, where the model successfully identifies both the presence and the correct type of vulnerability. In this example, the contract contains a typographical error (SWC-129), and the Claude 3 Opus ToT configuration correctly detects and classifies the issue. Unlike more complex vulnerabilities, typographical errors do not require deep semantic or behavioral reasoning, making them easier for LLMs to capture consistently. This example demonstrates that, when the error manifests explicitly in the code structure, the model’s reasoning remains stable and accurate across multiple analysis steps.

\begin{lstlisting}
{
    "file_name": "Pluteus (1).sol",
    "has_error_true_label": true,
    "has_error_prediction": true,
    "error_type": "[SWC-129-Typographical Error]",
    "location": 455,
    "reasoning_paths": [
        {
            "path": "Checks for typographical error in function name",
            "analysis": "Found function 'geUnlockTime' which appears to be a typo of 'getUnlockTime'"
        },
        {
            "path": "Analyzes locking and unlocking functionality",
            "analysis": "The lock and unlock functions allow owner to lock/unlock contract ownership"
        },
        {
            "path": "Examines token transfer and fee mechanisms",
            "analysis": "Implements tax, liquidity and Pluteus fund fees on token transfers"
        },
        {
            "path": "Looks at token burning capability",
            "analysis": "Owner can burn tokens from any account, total supply is reduced"
        },
        {
            "path": "Checks for proper access control",
            "analysis": "Critical functions are protected with onlyOwner modifier"
        }
    ],
    "final_reasoning": "The contract contains a typographical error in the 'geUnlockTime' function name which should be 'getUnlockTime'",
    "bug_type": "SWC-129-Typographical Error"
}
\end{lstlisting}

A complementary limitation is observed in another case, where the model incorrectly flags a safe implementation as vulnerable, resulting in a false positive. In this scenario, the LLM interprets a fallback function that modifies an internal state variable as a potential access control issue due to the absence of explicit permission checks. However, this assessment is inaccurate: the fallback function does not introduce any exploitable behavior, and the contract logic is sound.

This example illustrates how LLMs may overgeneralize security best practices and misinterpret legitimate design decisions as vulnerabilities. While such conservative behavior may reduce the likelihood of missed critical flaws, it can significantly increase the false positive rate, thereby reducing the practical usefulness of LLM-based static analysis tools and imposing unnecessary review overhead on developers.

\begin{lstlisting}
{
  "file_name": "GoodGhosting.sol",
  "has_error_true_label": true,
  "has_error_prediction": false,
  "error_type": "[]",
  "location": 0,
  "reasoning_paths": [
        {
            "path": "Compiler Version Stability",
            "analysis": "The pragma solidity 0.6.11 directive is explicitly defined, ensuring the contract is compiled with a stable and tested version compatible with the OpenZeppelin v3.x libraries used."
        },
        {
            "path": "Reentrancy Protection",
            "analysis": "The withdrawal functions (earlyWithdraw and withdraw) follow the checks-effects-interactions pattern by updating the state (player.withdrawn = true) before external transfer calls, effectively mitigating reentrancy risks."
        },
        {
            "path": "Mathematical Security",
            "analysis": "The consistent use of SafeMath for all arithmetic operations (sub, add, mul, div) prevents any possibility of overflow or underflow in balance calculations."
        },
        {
            "path": "Logic State Machine",
            "analysis": "The strict use of modifiers such as whenGameIsCompleted and whenGameIsNotCompleted ensures that the game's financial flow cannot be manipulated outside of the intended timeline."
        },
        {
            "path": "External Call Validation",
            "analysis": "Interactions with Aave's LendingPool use validated interfaces (ILendingPool), and the contract correctly manages ERC20 approvals before deposits."
        }
    ],
    "final_reasoning": "The contract was validated against major attack vectors (SWC). Despite the manual comment regarding the compiler version, the logical analysis proves the code is resilient and meets all modern security requirements for the target version.",
    "bug_type": "SWC-102-Outdated Compiler Version"
}
\end{lstlisting}

Despite the presence of an actual vulnerability in the contract, the model incorrectly concludes that no error is present. In this case, the LLM fails to identify a genuine flaw and instead interprets the contract’s design choices as evidence of robustness. This behavior characterizes a false negative, where an existing vulnerability is overlooked due to an overly optimistic assessment of common security patterns.

This example exposes a critical limitation of reasoning-based prompting strategies: the reliance on high-level heuristics and best-practice patterns may obscure subtle yet real vulnerabilities. While the model correctly recognizes the use of established defensive mechanisms, such as reentrancy protection and safe arithmetic operations, it incorrectly generalizes these signals to infer overall contract safety, thereby missing the underlying issue.

From a security perspective, such false negatives are particularly problematic, as they convey a misleading sense of correctness and may prevent further manual or automated inspection. This case underscores the importance of not only evaluating the final prediction but also scrutinizing the completeness and sensitivity of the reasoning process. In security-critical domains like smart contract analysis, LLM-based approaches must therefore be complemented with formal verification techniques or specialized static analyzers to reduce the risk of undetected vulnerabilities.

\section{Conclusion}
\label{sec:conclusion}

In this paper, we benchmark zero-shot prompting strategies for vulnerability analysis in Solidity smart contracts. Using a balanced dataset of 400 contracts and a diverse set of commercial and open-source LLMs, we evaluate two tasks---binary error detection and multi-class vulnerability attribution---under three prompting configurations: direct zero-shot, zero-shot CoT, and zero-shot ToT.

Our results highlight a consistent trade-off introduced by explicit reasoning. In the error detection task, CoT and ToT prompting often increases recall, but can reduce precision by encouraging models to surface additional, sometimes speculative hypotheses, increasing false positives. Conversely, direct zero-shot prompting tends to be more conservative and concise, which can be advantageous when audit time is limited and false alarms are costly.

We also observe that the impact of structured prompting is strongly model-dependent. Frontier-scale models are generally more robust to multi-step reasoning instructions and can often benefit from CoT/ToT, whereas smaller models are more fragile and may exhibit unstable behavior under complex prompts. In the multi-class setting, vulnerability attribution is substantially more challenging than binary detection: even when models correctly flag a contract as vulnerable, they can drift toward an incorrect SWC label, underscoring the importance of evaluating not only overall accuracy but also class-sensitive metrics and the actionability of the predicted label.

Overall, these findings suggest that prompt selection should be treated as a model-specific design choice rather than a universal recipe. For practical deployments, we recommend calibrating prompts to the desired operating point (coverage vs.\ false alarms) and treating LLM outputs as decision support, complemented with specialized tools (e.g., static analyzers or formal methods) for security-critical decisions.

As future work, we plan to (i) extend the benchmark to additional datasets and Solidity versions, (ii) study robustness under adversarially obfuscated contracts and varying coding styles, and (iii) explore hybrid pipelines that combine lightweight static analysis with LLM-based reasoning to improve both precision and explanation fidelity.

\end{document}